# Divide and Grow: Capturing Huge Diversity in Crowd Images with Incrementally Growing CNN

Deepak Babu Sam    Neeraj N Sajjan    R. Venkatesh Babu    Mukundhan Srinivasan
Video Analytics Lab, Indian Institute of Science    NVIDIA
Bangalore, India 560012    Bangalore, India 560045

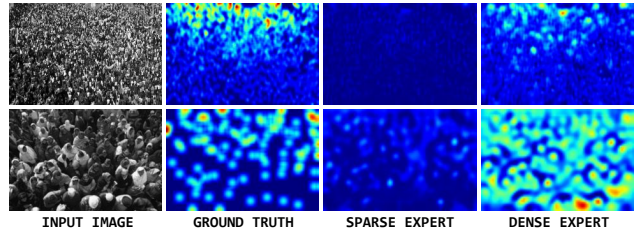

Figure 1. Predictions of a typical regressor fine-tuned for sparse or dense crowds. Models perform better on their own specialties.

## Abstract

*Automated counting of people in crowd images is a challenging task. The major difficulty stems from the large diversity in the way people appear in crowds. In fact, features available for crowd discrimination largely depend on the crowd density to the extent that people are only seen as blobs in a highly dense scene. We tackle this problem with a growing CNN which can progressively increase its capacity to account for the wide variability seen in crowd scenes. Our model starts from a base CNN density regressor, which is trained in equivalence on all types of crowd images. In order to adapt with the huge diversity, we create two child regressors which are exact copies of the base CNN. A differential training procedure divides the dataset into two clusters and fine-tunes the child networks on their respective specialties. Consequently, without any hand-crafted criteria for forming specialties, the child regressors become experts on certain types of crowds. The child networks are again split recursively, creating two experts at every division. This hierarchical training leads to a CNN tree, where the child regressors are more fine experts than any of their parents. The leaf nodes are taken as the final experts and a classifier network is then trained to predict the correct specialty for a given test image patch. The proposed model achieves higher count accuracy on major crowd datasets. Further, we analyse the characteristics of specialties mined automatically by our method.*

## 1. Introduction

Crowds are common in public places; be it the daily city traffic or some special gatherings, analysing crowds is becoming increasingly important both in terms of security as well as planning. Counting people in crowds, especially dense gatherings, is a difficult task even for humans. This is primarily because of the large diversity in the way people appear in crowded scenes. In highly dense crowds, people are only seen as blobs, while in less dense gatherings facial features may be visible. Hence, the visibility of features for crowd discrimination is related to the density of the crowd. Severe occlusion, pose changes and view-point variations further compound the problem. Head or body detection based methods fails to adapt to such huge diversity. As a result, most modern approaches solve counting by regression. With advent of deep learning, often Convolutional Neural Networks (CNN) are trained to predict crowd density map. Density maps represent count per unit area and provide spatial information. These models learn to map crowd features (head or shoulder patterns as they appear in crowds) to its density as opposed to detecting individuals with facial or body features.

Typical CNN density regressors [2, 10, 26] are optimized over an entire dataset containing images of all densities. And in many datasets, the density of crowd is not uniform. For example, in PartA of Shanghaitech dataset [26], the distribution is skewed with less number of dense crowd images than sparse ones. Consequently, performance of models varies widely across different categories of crowds. This usually results in over-estimating count for sparse images and under predicting for dense images as shown in [16].

One obvious solution to address this wide variability is having multiple regressors, each specialized for a particular type of crowd. This is illustrated in Figure 1 with predictions made by regressors fine-tuned for sparse and dense crowds. The experts perform well in their own specialties but worse in others. The major difficulty in such approaches is determining a criteria for creating experts. For Figure 1, we choose division based on density, but it can be based on other characteristics too. Even if the criteria is chosen, what would be the basis of division (how many people make a crowd dense or sparse)? These metrics are dataset as well as model dependent and need to be manually specified. In crowd counting, till now no principled method has been proposed to do this. Moreover, improper divisions can lead to suboptimal solutions. Learning experts automatically with classical mixture of expert [5] models do not work well in this scenario as shown by some works like [6].

The aim of this work is to introduce a principled methodology for creating experts, without any handcrafted dataset dependent criteria for specialization. Hence, we propose an Incrementally Growing CNN (IG-CNN) for crowd counting. IG-CNN starts from a base CNN density regressor which is trained on the entire dataset. Then we replicate the base CNN into two child networks by copying the weights of the parent. To make these child regressors specialized, we do differential training [13], where clustering of the dataset is done jointly with fine-tuning of the child networks (Section 3.4). In the next growing step, we replicate each of the child regressors again into two networks and perform differential training. This procedure is done recursively forming a hierarchical CNN tree where each node has two child nodes which are more specialized than their parent. At the end of the growing, a set of experts are created at the leaf nodes of the CNN tree. At test time, a classifier routes the input image patches to the appropriate expert regressors.

In a nutshell, this paper introduces the following:

- A hierarchical clustering method that jointly creates image clusters and a set of expert neural networks specialized on their respective clusters.

- A crowd counting system that can adapt and grow based on the complexity of the dataset.

## 2. Previous Work

**Crowd Counting:** There are numerous crowd counting methods in the literature which detect individual humans by identifying heads or other body structures [23, 20]. Such models primarily rely on hand-crafted features to train detectors. Stewart et al. [17] use a recurrent framework to sequentially detect people. All these detection methods fail in highly dense crowds as the discriminative features on which they can be trained are absent. Hence, regression based methods have gained traction. For example in [4], head detections along with features from interest points and Fourier analysis are used to regress crowd count.

With deep learning, CNN based regressors become popular and give better performance than classical models. The counting CNN introduced in [25], is trained by alternatively optimizing both crowd density loss as well as crowd count loss. The deep CNN employed by [19] directly regresses the crowd count instead of a density map. But such approaches prevent the CNN from acquiring good feature detectors and lead to lower performance than training for density map prediction. Walach et al. [18] train multiple CNNs where each network predicts corrections on the density map generated by the previous. In contrast, top-down feedback is leveraged by [12] to rectify initial detections of the bottom-up CNN regressor. Further works like [15, 16] supplement the main density regressors with high-level, low-level or both prior information. These priors are in the form of confidences predicted by a network trained to classify images based on density (sparse, dense etc.). This helps regressors to select specialized pathways based on the prior for generating density maps. But training of the classifiers require manual division based on density and is dataset dependent.

Approach by [10] makes use of multi-scale input and CNNs are trained for a particular scale. The last layer features of the networks are fused through learned fully connected layers. Though feeding multi-scale images can account for some scale variability, multi-columns CNNs are shown to be better. Boominathan et al. [2] propose a VGG based deep CNN along with a shallow CNN. The shallow CNN is designed to capture dense crowds while deep network is for sparse crowds. Multi-column network from [26] has three CNN columns, each having different receptive fields and hence can capture crowds at multiple scales. Features of the columns are fused through a learned $1 \times 1$ filter to generate the final density map. Since in these multi-column approaches the entire model is trained together, specialization gained among the columns need not be drastic. Sam et al. [13] address this issue by performing a differential training procedure to accentuate the specialization between CNN columns of varied architecture. But the model is limited by the availability of regressors with different architectures. In contrast, our method requires only one base regressor. More specific standard mixture of experts [5] based model is used in [6] for direct count regression, but performs worse than the hard switching mechanism of [13].

**Growing Networks:** The concept of a neural network that incrementally enlarges its capacity while learning has been there for a while. Several such Growing Neural Network models have been proposed in the literature [11, 3] for supervised as well as unsupervised learning. Few works like [9] grow a CNN progressively by adding new neurons in a data-dependent fashion. In the domain of transfer learning, [21] analyse different approaches for developmental networks which can increase its model capacity as and when new tasks are given. They explore adding new layers along the depth or width of the neural network.

**Specialization based Methods:** Expert specialization approaches like [24], increase classification performance by imposing coarse and fine class hierarchy onto a deep CNN. But this method requires coarse labels which is not required by method introduced in [1]. In a generalist-specialist setting, Ahmed et al. [1] jointly train specialty branches along with a generalist CNN which can classify the specialties. Specialty groups are formed such that they can be easily discriminated by the classifier. The algorithm proposed in [22] can learn a CNN tree where the nodes down the tree capture progressively fine-grained features. Our model also hierarchically grows a CNN tree, but it is employed only as a method to create a set of experts without any manually specified specialization criteria. In contrast to works like [22], the hierarchy is discarded in IG-CNN after training and only the networks at the leaf nodes of the CNN tree are kept. These leaf networks are finer experts and are selectively used at test time to evaluate inputs corresponding to their specialties.

## 3. Our Approach

### 3.1. Creating Experts with Hierarchical Differential Training

As motivated in Section 1, counting models have to handle severe diversity in the way people appear in crowds. We try to mitigate this issue with a set of expert regressors, each of which are specialized on one particular subset of the training data. Many previous works leveraging such specialization methods require the specialty information to be given either in the form of priors [15, 16] or indirectly as regressors with different architectures [13]. In this scenario, naive mixture of experts based methods are shown to perform subpar [6]. Hence, we design a model which does not require any specialty criteria to be manually specified for training experts and yet achieves better count estimates.

Our incrementally growing CNN or IG-CNN architecture is shown in Figure 2. IG-CNN training begins with a base CNN regressor, which is trained on the full dataset to estimate crowd density. To create specialties, a hierarchical training procedure is employed. Let $R_0$ represent the base CNN and $D_0 = \{X_{i=0}^N\}$ denotes the dataset of $N$ images on which the base regressor $R_0$ is trained. Initially, the base CNN is replicated into two networks $R_{00}$ and $R_{01}$. Now with differential training, $R_{00}$ and $R_{01}$ need to be made experts in separate specialties. The differential training procedure divides the dataset into two and fine-tunes the two regressors on their respective clusters. For a given image patch, only the regressor with the best count accuracy is trained. This way the *oracle loss* [7] of the set of regressors is minimized. *Oracle loss* is the minimum loss achievable if the correct regressor is chosen for all samples (see Section 3.4). We use a modified version of the differential training introduced in [13]. Our algorithm does not require regressors with different architectures as in [13] and also uses

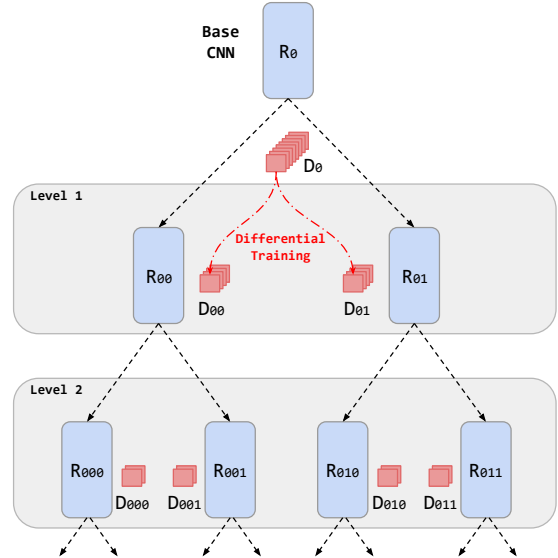

Figure 2. Hierarchical Differential Training in IG-CNN. Regressors are recursively replicated and specialized forming a CNN tree.

count loss function for fine-tuning.

After the first level of training, we have two expert regressors $R_{00}$ and $R_{01}$ along with their corresponding specialty subsets $D_{00}$ and $D_{01}$. Each of the networks $R_{00}$ and $R_{01}$, is replicated again to form respective child networks in the second iteration of growing. Regressors $R_{000}$ and $R_{001}$ will have same weights as of $R_{00}$ while $R_{01}$ is copied to $R_{010}$ and $R_{011}$. Differential training is performed on $R_{000}$ and $R_{001}$ with dataset $D_{00}$ only. Similarly, $D_{01}$ is used for fine-tuning $R_{010}$ and $R_{011}$. This makes sure that specialization acquired by the parent is propagated to its child networks. The two way replicating and specialization process is recursively continued forming a CNN tree. A child node in the tree is more specialized than any of their parent network. More deeper the tree, more finer the specialties with leaf nodes being the finest experts. Section 3.4 elucidates training algorithm in detail.

### 3.2. Growing CNN Architecture

The hierarchical differential training procedure results in the creation of a set of regressors at the leaf nodes of the CNN tree. All the regressors have the same architecture but give better performance on their specialties. Additionally, a classifier is trained for selecting the right expert for a given scene patch. Figure 3 shows the test time architecture of IG-CNN.

A neural network with five convolutional layers is used as the base CNN. Because of two pooling layers, the density prediction is at $\frac{1}{4}$th scale of the input image. All convolutional layers use ReLU activation function. This simple regressor is introduced in [26] and delivers reasonable perfor-mance. But it is to be noted that our training methodology is generic and is not limited by any particular base CNN. For expert classifier, we use a modified VGG-16 [14] network. Features of the last convolution layer of VGG-16 are reduced via global average pooling followed by two fully connected layers and softmax prediction at the end. The number of units in the last fully connected layer depends on how many expert regressors are generated with the hierarchical training.

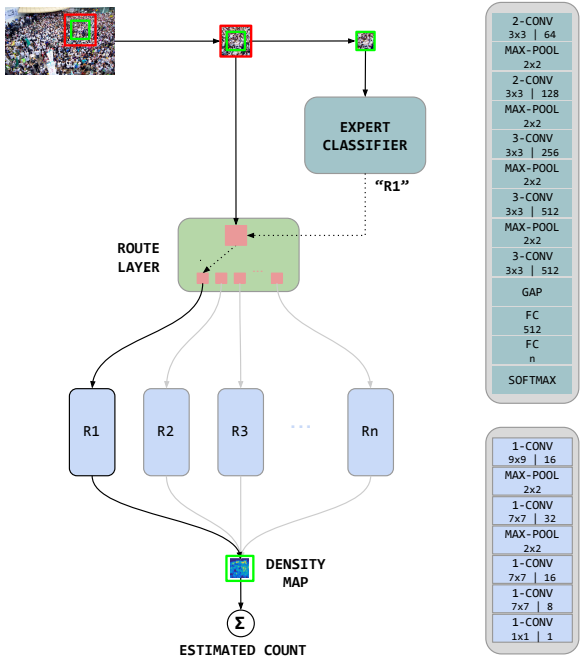

Figure 3. Test time architecture of IG-CNN. The expert classifier routes crowd patches to the appropriate specialized regressor.

During testing, we take overlapping patches from the image. Each patch extracted is of the size $P_W \times P_H$. A region of interest (RoI) of size $R_W \times R_H$ is defined for the patch on which the CNN regressor predicts the crowd density. Area outside the RoI acts as context and aids in better regression. The RoI is slided over the entire image with an overlap. The predictions of the overlapping areas are averaged to get the final density. Characteristics of crowd in the RoI is assumed to be constant. For IG-CNN hierarchical differential training, patches are extracted at random locations from images and the loss is computed only on the RoI. The expert classifier also uses the RoI part of a patch to select the suitable regressor. Typical patch size is $224 \times 224$ with an RoI size of $80 \times 80$.

### 3.3. Pretraining of Base CNN

The base CNN is trained on the entire dataset to regress crowd density map. The network is trained by backpropagating $l_2$ loss between the predicted and the ground truth density maps. There are numerous methods in the literature to generate ground truth density maps from head annotations available with the datasets [25, 26]. The most widely used method is to place a Gaussian at every head location. This helps CNN to train better as otherwise it would have to predict exactly at the annotation point. The spread of the Gaussian is an hyper-parameter and depends on the dataset. Since Gaussian blurring of each annotation sums to one, the crowd count can be obtained by summing the density map.

Let $M_{X_i}(x; \Theta)$ denote the density map predicted by the CNN regressor and $M_{X_i}^{GT}(x)$ be the ground truth for image $X_i$. Then, the $l_2$ loss function is defined as

$$L_{l_2}(\Theta) = \frac{1}{2N} \sum_{i=1}^{N} \|M_{X_i}(x; \Theta) - M_{X_i}^{GT}(x)\|_2^2, \quad (1)$$

where $\Theta$ refers to the trainable parameters of the CNN and $N$ is the total number of training samples. The parameters $\Theta$ are found by optimizing $L_{l_2}$ using standard stochastic gradient descent (SGD) with momentum. Even though our objective is to minimize the count error, $l_2$ loss acts as proxy for the count loss. Reduction in $l_2$ distance implicitly lowers the count error between the prediction and ground truth. For pretraining, we crop patches at different locations from every image and apply flip augmentation.

### 3.4. Training Algorithm for IG-CNN

The overall training procedure of IG-CNN is depicted in Algorithm 1. The first step is the pretraining of the base CNN $R_0$. For any regressor, the final metric of performance is the count error which needs to be minimized. Count predicted by a regressor $R$ for an image $X_i$ is computed as $C_{X_i}^R = \sum_x M_{X_i}(x; \Theta_R)$, where its ground truth count is $C_{X_i}^{GT} = \sum_x M_{X_i}^{GT}(x)$. The count error for regressor $R$ on image $X_i$ is the absolute difference of predicted and actual count or mathematically, $E_{X_i}(R) = |C_{X_i}^R - C_{X_i}^{GT}|$.

After pretraining of the base CNN, a CNN tree is progressively built where each node represents a regressor finetuned on a subset of the dataset. This is done by replicating each regressor at the tree leaves into two and specializing the child networks with differential training. At any node $m$, let $R_{m0}$ and $R_{m1}$ be the child regressors and $D_m$ be the subset of dataset available for the node. Now $R_{m0}$ and $R_{m1}$ need to be made experts in the specialty sets $D_{m0}$ and $D_{m1}$ respectively. But we have neither the specialty sets nor the expert regressors. Differential training allows to jointly obtain the specialties and the expert regressors by minimizing the oracle count error. The oracle count error for patch $X_i$ is $E_{X_i}^{oracle} = \min_{R \in [R_{m0}, R_{m1}]} |C_{X_i}^R - C_{X_i}^{GT}|$, the minimum of the count errors obtained by the two regressors. The basic idea is to evaluate both the regressors on a particular image patch and fine-tune only the one giving lesser count error. Choose the best regressor by $r_{X_i}^{best} = \operatorname*{argmin}_{R \in [R_{m0}, R_{m1}]} |C_{X_i}^R - C_{X_i}^{GT}|$.

Note that when count predictions by both networks are same, which mostly happens at the start of the training, the first regressor is chosen to break the tie. This makes sure that the differentiation between the networks builds up progressively. By selectively fine-tuning $R_{m0}$ and $R_{m1}$ based on its performance on the training patches, the regressors become more and more specialized in two groups $D_{m0}$ and $D_{m1}$. These specialty subsets might be skewed and completely depends on the dataset as well as the regressors. The fine-tuning is done with lower learning rate ($10^{-6}$) and continue till validation accuracy stops improving.

Unlike differential training in [13], count loss is used instead of $l_2$ loss for fine-tuning regressors. We define the count loss as,

$$L_C(\Theta) = \frac{\lambda}{2N} \sum_{i=1}^{N} (C_{X_i} - C_{X_i}^{GT})^2. \quad (2)$$

Here constant $\lambda$ is used to check the magnitude of the loss. For all experiments, $\lambda$ is set as $10^{-2}$. Since the CNN is pretrained with $l_2$ loss, it has good initial features and fine-tuning with count loss provides complementary information. This is found to give better clustering and more accurate count estimation.

Differential training minimizes oracle error over the training set. This count error is achievable only if there is an oracle to classify a test patch to the correct regressor. But the ability of a classifier to achieve high results in determining the specialty depends on the quality of the specialization. If the expert specialties do not have any generalizable features, the performance might decay on the test set.

The leaf regressors ($R_{leaf}$) at a particular level of growth are experts on specific specialties. As shown in Figure 3, at test time, a classifier selects the right expert regressor for the image patch. The classifier is trained on the labels obtained from the expert regressors. For a given image patch $X_i$, the corresponding label is attributed to the regressor with minimum count error, $R_{X_i}^{best} = \underset{R \in R_{leaf}}{\operatorname{argmin}} |C_{X_i}^R - C_{X_i}^{GT}|$. As the samples per expert specialty need not be uniform, class balancing is done before training the classifier.

At every increment of the growing process, regressors at the leaf nodes of the CNN tree are split and new expert regressors are created. We monitor the *Oracle MAE* and *Actual MAE* for the leaf regressors over a validation set. Mean Absolute Error or MAE is the performance metric used for counting models (see Section 4.1). While *Oracle MAE* indicates the count error incurred when right expert is always chosen for regression, *Actual MAE* is obtained with the expert classifier. Note that the validation set is randomly sampled from the training images and is fixed across entire training procedure (irrespective of tree level). The hierarchical tree splitting is stopped when the *Actual MAE* on validation set is not improving (see Table 4).

---

**input** : Dataset $D_0 = \{X_i, M_{X_i}^{GT}\}_{i=1}^{N}$ (image & map)
**output:** Parameters $\{\Theta_r\}$ of experts and classifier $\Theta_c$
Random initialize $\Theta_0$ for base CNN $R_0$;
Pretrain $R_0$;
$R_{leaf} = \{R_0\}$;
$D_{leaf}[R_0] = D_0$;
/* Hierarchical Differential Training */
**for** *l = 0 to max_tree_depth* **do**
    /* Replicate $R$ twice */
    **for** *R in $R_{leaf}$* **do**
        $R_{child}[R] = \{R, R\}$;
    **end**
    /* Differential Training */
    /* $R$ predicts count $C_i^R$ while $C_i^{GT}$ is the actual */
    **for** *i = 1 to max_iterations* **do**
        **for** *$R_l$ in $R_{leaf}$* **do**
            **for** *X, M in $D_{leaf}[R_l]$* **do**
                $r = \underset{R \in R_{child}[R_l]}{\operatorname{argmin}} |C_X^R - C_X^{GT}|$;
                Fine-tune $R_r$ with $X$ to update $\Theta_r$;
            **end**
        **end**
        Break if validation *Oracle MAE* stagnates;
    **end**
    /* Dataset division for leaf regressors */
    $D_{leaf} = []$;
    **for** *X, M in $D_0$* **do**
        **for** *$R_l$ in $R_{leaf}$* **do**
            $r = \underset{R \in R_{child}[R_l]}{\operatorname{argmin}} |C_X^R - C_X^{GT}|$;
            Add $(X, M)$ to $D_{leaf}[r]$;
        **end**
    **end**
    /* Training of Expert Classifier */
    Initialize $\Theta_c$ with VGG-16 weights;
    **for** *X, M in $D_0$* **do**
        $r = \underset{R \in R_{leaf}}{\operatorname{argmin}} |C_X^R - C_X^{GT}|$;
        Add $(X, r)$ to $D_c$;
    **end**
    Train classifier with $D_c$ and update $\Theta_c$;
    Break if validation *Actual MAE* stagnates;
**end**

**Algorithm 1:** IG-CNN training algorithm.

## 4. Experiments

### 4.1. Evaluation Scheme

We benchmark our IG-CNN model on three crowd counting datasets. For a given test image, patches are extracted and evaluated by the expert classifier to route them to the regressors specialized for the specific crowd types.

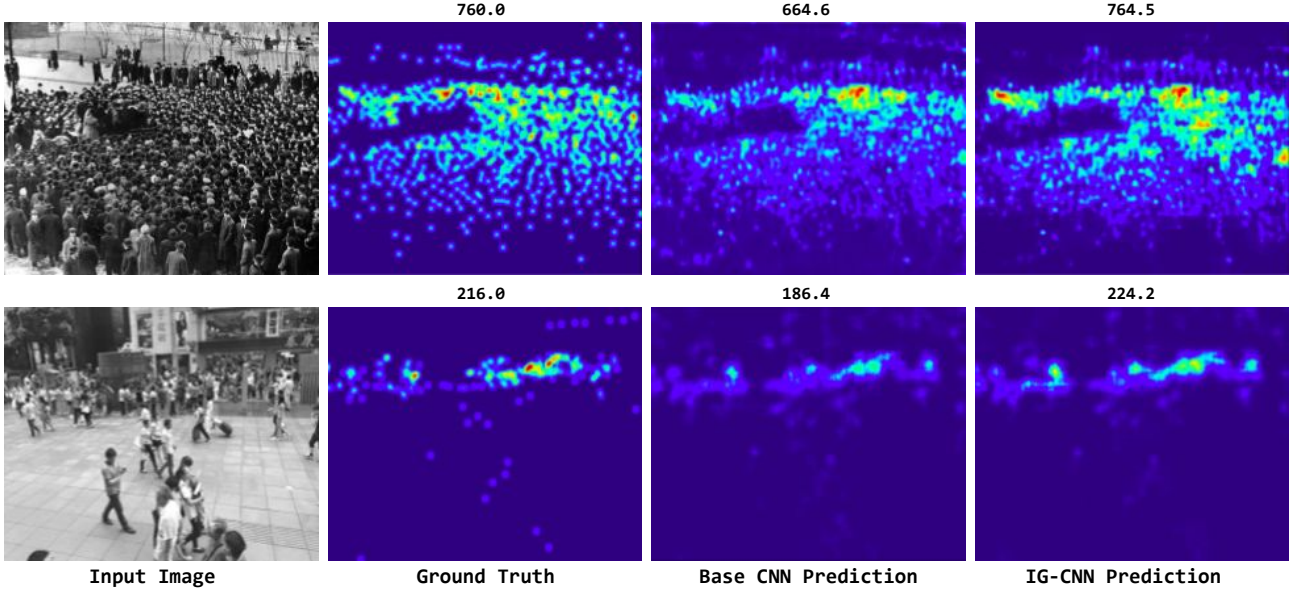

Figure 4. Predictions made by IG-CNN on images of Shanghaitech dataset [26].

Two metrics are commonly used to evaluate crowd counting models. MAE or Mean Absolute Error is the most important metric and indicates the count accuracy of the model. If the count predicted by the model is $C_{X_i}$ for an actual count of $C_{X_i}^{GT}$, then MAE is defined as

$$\text{MAE} = \frac{1}{N} \sum_{i=1}^{N} |C_{X_i} - C_{X_i}^{GT}|, \qquad (3)$$

where the test set has $N$ images. The second metric is the Mean squared error or MSE to measure the variance of count estimation. MSE is given by,

$$\text{MSE} = \sqrt{\frac{1}{N} \sum_{i=1}^{N} (C_{X_i} - C_{X_i}^{GT})^2}. \qquad (4)$$

### 4.2. Shanghaitech dataset

The Shanghaitech dataset [26] is partitioned into two sets, namely Part_A and Part_B. The sets are quite distinctive with Part_A images sourced from the Internet and have crowd counts ranging from 33 to 3139. In contrast, the crowds in Part_B are sparser with counts varying from 9 to 578 and are captured from streets of Shanghai. Part_A has 482 images, of which 300 images are used for training and the rest are used for testing. Similarly, the 716 images of Part_B are partitioned into chunks of 400 training and 316 testing images. Gaussian kernels with fixed sigma are used to generate the ground truth density maps.

For both Part_A and Part_B, we grow IG-CNN to 3 levels resulting in 8 expert regressors. Table 1 tabulates the performance metrics for IG-CNN on the dataset along with that of other models. It can be observed that IG-CNN outperforms all other methods in Part_B by a significant margin both in terms of MAE and MSE. IG-CNN achieves better count accuracy in Part_A as well. Though our model narrowly outperforms CP-CNN [16], it is to be noted that the authors of CP-CNN use adversarial training to boost their base performance from 76.1 to 73.6. Figure 4 shows density maps predicted by IG-CNN and the base CNN along with the corresponding ground truths. The predicted density maps closely resemble the ground truth as well as have accurate count estimates. This demonstrates the ability of IG-CNN to better capture the crowd density.

### 4.3. UCF_CC_50 dataset

UCF_CC_50 dataset introduced by [4], serves to be a hard challenge because of the relatively small size of the dataset and the large variability of crowd density in the images. The crowd count in 50 images of the dataset vary considerably from 94 to 4543. The ground truth density maps are created with fixed variance Gaussian kernel. We follow 5-fold cross-validation protocol adopted by [4] to evaluate the model on the dataset.

|  | Part_A | | Part_B | |
|---|---|---|---|---|
| Method | MAE | MSE | MAE | MSE |
| Zhang [25] | 181.8 | 277.7 | 32.0 | 49.8 |
| MCNN [26] | 110.2 | 173.2 | 26.4 | 41.3 |
| SCNN [13] | 90.4 | 135.0 | 21.6 | 33.4 |
| TDF-CNN [12] | 97.5 | 145.1 | 20.7 | 32.8 |
| CP-CNN [16] | 73.6 | **106.4** | 20.1 | 30.1 |
| IG-CNN | **72.5** | 118.2 | **13.6** | **21.1** |

Table 1. Performance of IG-CNN on Part_A and Part_B of Shanghaitech dataset [26]. IG-CNN outperforms other methods in MAE.

| Method | Scene1 | Scene2 | Scene3 | Scene4 | Scene5 | Average |
|---|---|---|---|---|---|---|
| Zhang et al. [25] | 9.8 | **14.1** | 14.3 | 22.2 | **3.7** | 12.9 |
| MCNN [26] | 3.4 | 20.6 | 12.9 | 13.0 | 8.1 | 11.6 |
| SCNN [13] | 4.4 | 15.7 | **10.0** | 11.0 | 5.9 | 9.4 |
| CP-CNN [16] | 2.9 | 14.7 | 10.5 | **10.4** | 5.8 | **8.9** |
| IG-CNN | **2.6** | 16.1 | 10.15 | 20.2 | 7.6 | 11.3 |

Table 2. MAEs obtained by models for the 5 test scenes of WorldExpo'10 dataset [25].

IG-CNN hierarchical growing is done for two levels, creating 4 expert regressor on UCF_CC_50 dataset. It can be seen from Table 3 that IG-CNN has the lowest count error. Despite being a challenging dataset, our model delivers an improvement of 4.4 points in MAE and has comparable performance in MSE metric also.

| Method | MAE | MSE |
|---|---|---|
| Lempitsky et al. [8] | 493.4 | 487.1 |
| Idrees et al. [4] | 419.5 | 541.6 |
| Zhang et al. [25] | 467.0 | 498.5 |
| CrowdNet [2] | 452.5 | - |
| MCNN [26] | 377.6 | 509.1 |
| Hydra2s [10] | 333.7 | 425.3 |
| SCNN [13] | 318.1 | 439.2 |
| Cascaded-MTL [15] | 322.8 | 397.9 |
| CP-CNN [16] | 295.8 | **320.9** |
| IG-CNN | **291.4** | 349.4 |

Table 3. Comparison of IG-CNN with other methods on UCF_CC_50 dataset [4]. Our model gives lower error than other methods.

### 4.4. WorldExpo'10 dataset

The World Expo'10 dataset introduced by [25] is a large dataset containing 1132 video sequences captured by 108 surveillance cameras covering a set of scenes. On an average, each image has 50 people in it. The dataset is divided into two parts for training and testing. Training data consists of 3380 frames from 103 different scenes, whereas the test data comprising of 5 different scenes has a total of 600 frames. Along with the images, the authors have also provided the Region of Interest (RoI) and the perspective maps. The RoIs are used during training to backpropagate only in those regions. Also, only the prediction in the RoI is used to report crowd count while testing.

Table 2 lists the performance of all major methods. IG-CNN is grown only just one level with two experts. World Expo'10 dataset proves to be extremely challenging for our model due to the sparse nature of the crowd with the lack of significant variability in crowd density. This affects the ability of our model to generate experts catering to different crowd types. Despite these limitations, our model shows comparable performance with respect to other models.

## 5. Analysis and Ablations

### 5.1. Effect of Growing

In this section, we study the effect of the hierarchical CNN tree growing on the oracle accuracy and the final accuracy at test time. All ablations are performed on Part_A of the Shanghaitech dataset [26] as it is sufficiently large and has high variation in crowd density. Table 4 lists count errors for the base CNN along with that of the IG-CNN at different levels of growth. It also shows for each level, the classifier accuracy and Oracle MAE. This oracle error is the MAE that the model would achieve if the expert classifier is 100% accurate. There is significant improvement of MAE for IG-CNN at higher levels than the base CNN but saturates after level 3. Although the oracle error decreases down drastically with each increment of the growth, the expert classifier is unable to keep up and causes more switching error as evident from the Table 4. This is primarily due to the reduction in number of training samples per expert regressor at higher tree levels. For example at level 2, the distribution of samples for the four regressors is so skewed that one of the expert gets only 2.9% of the total test patches. This is more severe for level 3 with only 0.5% for the expert with the least number of samples and the corresponding class wise classifier accuracy is just around 2%. The number of samples for some of the regressors are so small that the classifier is unable to generalize significant discriminative features for the specialties. We also show in Table 4, the performance when the regressor with the least number of samples is not split, leading to an unbalanced tree. In this way, there are only 7 expert regressors at level 3 instead of 8 experts. The MAE in this case is comparable to IG-CNN at level 3, but higher.

### 5.2. Expert Specialty Characteristics

It is important to shed more light on the specialization process involved in the IG-CNN training. Hence, we analyse the features of specialty groups automatically inferred in the hierarchical differential training. The specialty groupings might be based on some latent features such that the oracle error is minimized. But are there observable characteristics based on crowd types? Crowd density could be one of the factors for specialization. Number of people in a image patch is a proxy for crowd density. We compute the distribution of crowd counts on the specialty subsets of the expert regressors (see Section 3.4 for classifier label creation). For the experiment, the test set of Part_A Shanghaitech dataset [26] is used. Figure 5 indicates a possible clustering of crowd patches based on count. Note that patches with few people go to one regressor while more denser ones get distributed across the other experts. This multichotomy observed in the specialties reinforces the fact that IG-CNN training creates experts based on certain latent factors. Some of the factors could be correlated with crowd density as density variation accounts for much of the variability seen in crowd images.

| Method | Oracle MAE | Actual MAE | Classifier Accuracy |
|---|---|---|---|
| Base CNN | - | 120.9 | - |
| 2 Experts (Level 1) | 38.1 | 115.3 | 77.2 |
| 4 Experts (Level 2) | 17.8 | 80.3 | 62.3 |
| 7 Experts (Level 3) | 11.4 | 78.1 | 45.7 |
| 8 Experts (Level 3) | 8.5 | 72.5 | 45.5 |
| 16 Experts (Level 4) | 4.4 | 74.6 | 21.8 |

Table 4. Effect of hierarchical growth of IG-CNN on Part_A of Shanghaitech [26] dataset. Though the oracle loss is steadily decreasing with depth, classifier error is increasing leading to higher MAE at test time.

### 5.3. Hierarchical Training Vs Baseline Methods

In short, IG-CNN training mines latent specialties hierarchically and creates a set of expert regressors. Here we compare this methodology with other similar methods. The standard mixture of experts (MoE) approach uses a gating network to weigh the output of the set of regressors. In the same setting as that of IG-CNN, we use VGG-16 classifier as gating CNN to output softmax confidences. The 8 regressors are initialized with base CNN weights and their outputs are multiplied by the classifier confidences. Table 5 shows MAE numbers for MoE and is clearly inferior to IG-CNN. MoE is unable to bring significant specialization among the regressors.

We also compare with differential training introduced in [13]. Instead of performing hierarchical training, N-way differential training is done on the set of regressors as in [13]. For this ablation, we use four and eight regressors which are exact copies of the base CNN, is comparable to IG-CNN with the same number of experts. The oracle loss of the expert set is minimized by selectively fine-tuning the best regressor for the given training sample. It can be observed from Table 5, that the oracle MAE is lower for

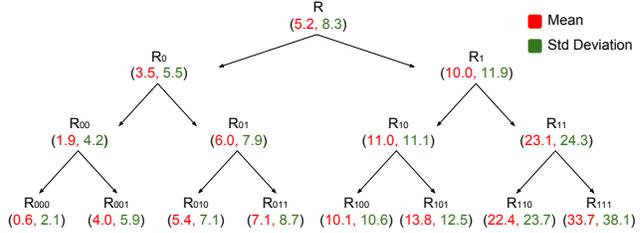

Figure 5. Mean and standard deviation of crowd count distribution preferred by expert regressors at different hierarchies of IG-CNN. Computed on patches from Shanghaitech [26] Part_A test set.

IG-CNN than that of N-way differentially trained model. In fact, the final performance with the expert classifier is also inferior in the case of N-way differential training. This emphasizes that the hierarchical training creates specialties with better discriminative features.

| Method | Oracle MAE | Actual MAE |
|---|---|---|
| Mixture of Experts | - | 281.8 |
| 4-way Differential Training | 20.6 | 99.0 |
| 8-way Differential Training | 9.9 | 75.1 |
| IG-CNN (Level 3) | **8.5** | **72.5** |

Table 5. Comparison of IG-CNN with other specialization based methods on Part_A of Shanghaitech [26] dataset. IG-CNN outperforms other architectures.

## 6. Conclusion

We address the problem of better capturing large diversity seen in crowd scenes for accurate regression of crowd density. The proposed model, IG-CNN iteratively expands its model capacity based on the complexity of the training data. IG-CNN starts growing from a base CNN, which is trained to regress crowd density. The base CNN is replicated into two child regressors, each of which are imposed specialization with differential training and recursively divided again forming a CNN tree. The regressors at the leaf nodes of the tree are finer experts on certain specialties mined without any manually specified criteria. An expert classifier predicts the right expert for a given test patch. We evaluate on standard benchmarks and show better performance for the model. Additionally, analysis of the specialties created by IG-CNN reveals correlation with observable crowd characteristics such as crowd density.

## 7. Acknowledgements

This work is supported by SERB, Department of Science and Technology (DST), Government of India (Proj No. SB/S3/EECE/0127/2015). We also acknowledge NVIDIA AI Tech Centre (NVAITC) for their assistance.